\documentclass[12pt]{article}

\usepackage{newtxtext,newtxmath}

\usepackage{graphicx}
\usepackage{subcaption}
\usepackage{cuted}
\usepackage{graphicx}
\usepackage{hyperref}
\usepackage{breakurl}
\usepackage{amsmath}

\usepackage{graphicx}

\usepackage[letterpaper,margin=1in]{geometry}

\renewenvironment{abstract}
	{\quotation}
	{\endquotation}

\date{}

\makeatletter
\renewcommand{\fnum@figure}{\textbf{Figure \thefigure}}
\renewcommand{\fnum@table}{\textbf{Table \thetable}}
\makeatother

\usepackage{scicite}

\usepackage{url}

\def\scititle{
Enhancing Reinforcement learning in 3-Dimensional Hydrophobic-Polar Protein Folding Model with Attention-based layers
}
\title{\bfseries \boldmath \scititle}

\author{
	Peizheng~Liu$^{1\ast}$,
	Hitoshi~Iba$^{1}$\\
	\small$^{1}$Faculty of Engineering, The University of Tokyo\\
	\small$^\ast$Corresponding author. Email: liupp55@g.ecc.u-tokyo.ac.jp
}

\begin{document} 

% Insert the title and author list
\maketitle

% Abstract, in bold
% There are strict length limits, and not all formats have abstracts.
% Consult the journal instructions to authors for details.
% Do not cite any references in the abstract.
\begin{abstract} \bfseries \boldmath
% Start with one or two sentences of background
Transformer-based architectures have recently propelled advances in sequence modeling across domains, but their application to the hydrophobic-hydrophilic (H-P) model for protein folding remains relatively unexplored. In this work, we adapt a Deep Q-Network (DQN) integrated with attention mechanisms (Transformers) to address the 3D H-P protein folding problem. Our system formulates folding decisions as a self-avoiding walk in a reinforced environment, and employs a specialized reward function based on favorable hydrophobic interactions. To improve performance, the method incorporates validity check including symmetry-breaking constraints, dueling and double Q-learning, and prioritized replay to focus learning on critical transitions. Experimental evaluations on standard benchmark sequences demonstrate that our approach achieves several known best solutions for shorter sequences, and obtains near-optimal results for longer chains. This study underscores the promise of attention-based reinforcement learning for protein folding, and created a prototype of Transformer-based Q-network structure for 3-dimensional lattice models.

\end{abstract}

\section{Introduction}
% The first paragraph of any Science paper does NOT have a heading
% Nor is it indented
\noindent
\textbf{H-P model} has been considered as a simplified model for protein structure prediction. However, optimizing the structure of H-P model still requires efficient algorithms due to the large solution space. Determining the optimal structure of proteins under the hydrophobic–hydrophilic (HP) model has been rigorously shown to be NP-complete\cite{berger1998protein}, highlighting the necessity for powerful heuristic or approximation methods in lieu of brute-force searches. 

Among heuristic approaches, Monte Carlo simulations are particularly popular and exhibit a wide range of implementations\cite{li2011monte}\cite{liang2001evolutionary}. In particular, the pruned-enriched Rosenbluth method (PERM)—a sequential Monte Carlo algorithm with resampling—has emerged as a strong performer for HP folding, producing several of the best-known solutions reported to date \cite{hsu2011review}. Likewise, replica exchange Monte Carlo (REMC) has also demonstrated top-level performance \cite{thachuk2007replica}.

Beyond Monte Carlo–based methods, various evolutionary algorithms have also shown efficacy in this domain. For instance, ant colony optimization (ACO) has been successfully applied to both two-dimensional and three-dimensional formulations of the HP model, yielding competitive results \cite{shmygelska2005ant}. Other well-known evolutionary and swarm-based methods—including genetic algorithms\cite{khimasia1997protein}\cite{lin2011protein} and particle swarm optimization\cite{lin2011protein}—have been tested on HP folding problems, also achieving effective approximations.

Recently, machine learning approaches have also been employed to tackle HP folding. Among these approaches, deep reinforcement learning (DRL) has emerged as a promising method due to its ability to handle sequential decision-making tasks in high-dimensional and complex environments. One pioneering study applying RL to the HP folding task was "FoldingZero" by Li et al. in 2018\cite{li2018foldingzero}. That work utilized a deep convolutional neural network (CNN) integrated with a Residual Upper Confidence Tree (R-UCT) approach, all trained using an RL algorithm. Further advancements in the application of RL to HP folding were explored by Yang et al.\cite{yang2023applying}, who evaluated several foundational approaches from recent RL studies. These included the use of Long Short-Term Memory (LSTM) networks\cite{graves2012long}, replay memory, and the Double Deep Q-Network (Double DQN)\cite{van2016deep}. 
\newline

% Below to be redefined
\noindent
\textbf{Transformer models}, first introduced by Vaswani et al. in the landmark paper "Attention Is All You Need" (2017)\cite{vaswani2017attention}, have emerged as a pivotal innovation in the field of machine learning. At their core, Transformers leverage the attention mechanism, a method that allows the model to assign varying levels of importance to different parts of the input data. This approach addresses key limitations of earlier sequence-based architectures such as recurrent neural networks (RNNs) and long short-term memory networks (LSTMs), which are constrained by their sequential processing nature and difficulties in capturing long-range dependencies\cite{vaswani2017attention}.

In this study, we extend the application of the transformer models to address the H-P model. While deep reinforcement learning techniques, such as Deep Q-Networks (DQNs) combined with RNNs or LSTMs, have been explored in H-P model research, transformer architectures remain underexplored. This gap presents an opportunity to investigate the potential advantages of transformers in modeling the unique characteristics of the H-P problem.

The H-P model encapsulates complex, non-sequential dependencies between folding steps driven by the spatial constraints. Traditional sequential models, such as RNN-LSTM structure, often struggle to capture these intricate dependencies due to their reliance on step-by-step processing. In contrast, the transformer model, with its self-attention mechanism, is inherently able to model such non-local interactions by attending to all relevant steps in parallel. This capability aligns well with the demands of H-P folding, which we aim to leverage in this study.

%%%%%%%%%%%%%%%% maybe add a figure for a non-sequential dependency

\section{Methodology}

To address the H-P protein folding problem, we use a self-avoiding walk (SAW) process. This environment emulates the folding process by enabling folding actions to be applied, and it generates observations encapsulating the resulting structural, serving as a dynamic platform for simulating and analyzing folding pathways.

Within this framework, a folding agent is employed to interact with the environment. The agent, powered by a Deep Q-Network (DQN) with attention-based layers, makes sequential decisions that guide the folding process. Through iterative interaction, the agent learns to optimize folding strategies by leveraging feedback from the environment, ultimately improving its decision-making capabilities.

\begin{figure}
    \centering
    \includegraphics[width=0.8\linewidth]{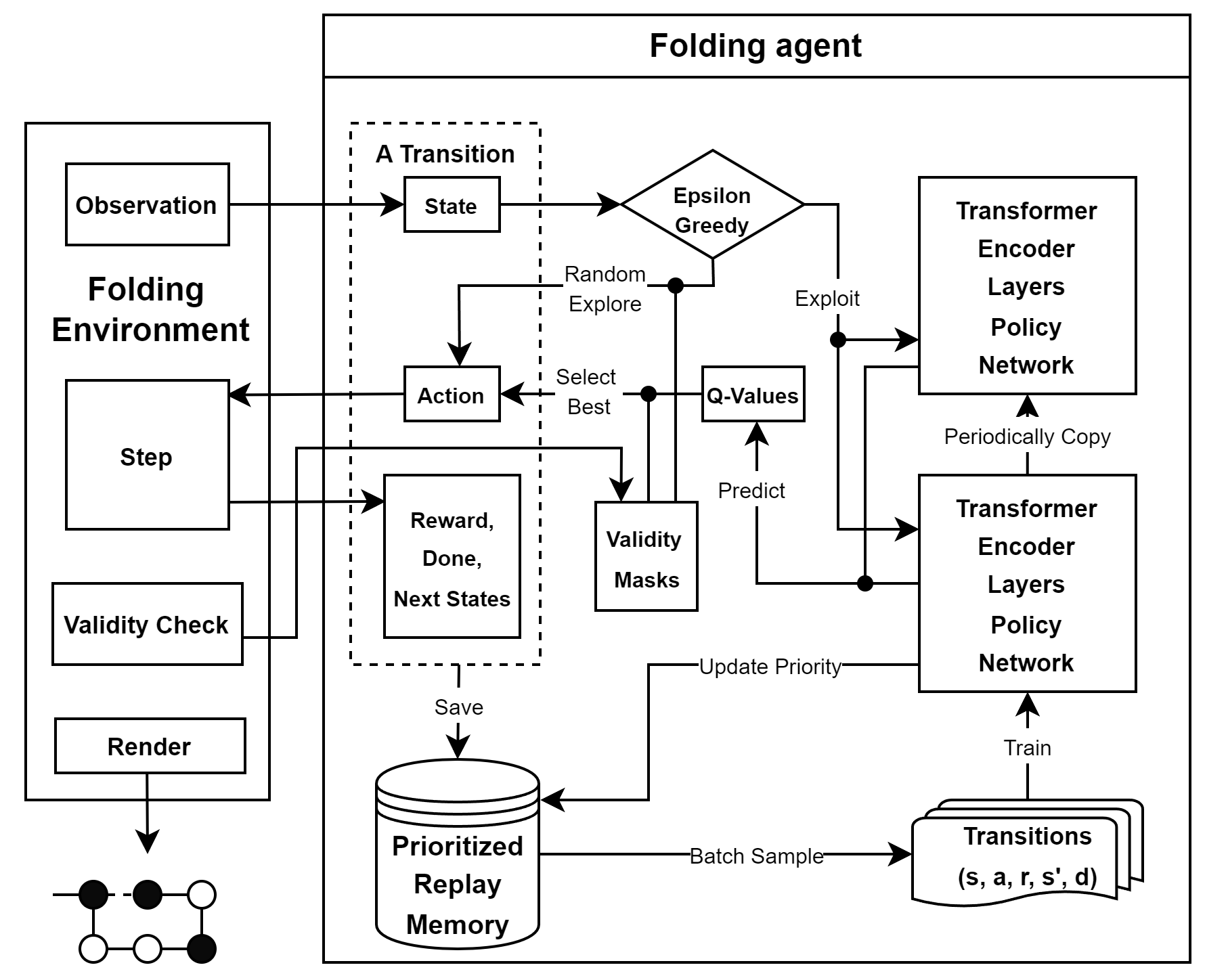}
    \caption{\textbf{Overall Training Data Flow.} The process begins with an empty folding environment, where the initial state \( s_0 \) is obtained by the agent. The agent iteratively follows a state-action-reward cycle until an episode terminates. After termination, the reward is backpropagated through each step and the environment is reset empty for the next episode. Training of the policy network is performed on a batch of samples from the replay memory after each step. The target network is periodically updated by copying the parameters of the policy network.}
    \label{fig:overall_training_figure}
\end{figure}

A brief overview of the entire training data flow is shown in Figure~\ref{fig:overall_training_figure}. The detailed methodology, including the environment's design, the structure of the neural network of the folding agent, and the training protocol, will be described in the following sections.

%%%%%%%%% figure of the agent and environment

\subsection{H-P Model Folding Environment}

The protein folding environment is constructed on a three-dimensional cubic lattice implemented with Python lists to represent the spatial arrangement of protein structures. Within this lattice, amino acids are positioned at specific, discrete coordinates, ensuring that each residue occupies a unique and fixed location in space. The input sequence of amino acids is encoded as a string of 'H' (hydrophobic) and 'P' (polar) characters. Placements and observations interact with lattice data through class functions defined in the standard format of the \textsc{gymnasium.Env} class from the OpenAI Gymnasium framework.

The OpenAI Gymnasium framework provides a standardized interface for defining and interacting with reinforcement learning environments. Each environment is implemented as a Python class that defines an \textit{action space} and an \textit{observation space}. The \textit{action space} specifies the set of all possible actions an agent can take, while the \textit{observation space} describes the format and bounds of the data received by the agent. This framework allows agents to interact with the environment in discrete time steps by selecting actions, observing resulting states, and receiving rewards, facilitating the design and evaluation of reinforcement learning tasks.

The following subsections elucidate the core components and operational mechanics of the folding environment.

\subsubsection{Amino Acid Placement Action}

The initial two amino acids in the sequence are fixed at predefined coordinates $(0, 0, 0)$ and $(1, 0, 0)$, establishing the initial direction for the folding process. Subsequent amino acids are introduced sequentially, with their placement governed by relative directional instructions derived from discrete actions. 

The environment utilizes a discrete \textit{action space} comprising five possible moves: continuing in the current direction (forward), deviating upward, deviating downward, deviating to the left, or deviating to the right. These directional shifts are represented as unit vectors in the three-dimensional grid. 

In each simulation, an \textit{episode} refers to a sequence of actions that ultimately leads to an end state (i.e., a done flag is returned). In this experiment, an end state occurs either when every amino acid has been placed by repeating the five actions described above, or when the agent selects an illegal action (as detailed in Section \ref{sec:invalid_move}).

\subsubsection{Structure Observation}

The environment's \textit{observation space} is defined as a continuous three-dimensional vector encapsulating the positions of each amino acid and their respective types. Specifically, the observation of a structure with length $l$ is a flattened array of shape $(l \times 5)$, where each amino acid is represented by its $x$, $y$, $z$ coordinates, a numerical encoding of its type (1 for 'H' and 0 for 'P'), and a normalized index value of the peptide sequence. Unplaced amino acids are denoted with placeholder values to maintain a consistent observation size. 

Moreover, the environment restricts all coordinates to lie within a cubic bounding region of side length $l$, centered at the origin. In other words, if $l$ is the sequence length, then each amino acid's $x$, $y$, and $z$ coordinates are constrained such that $\left| x \right|, \left| y \right|, \left| z \right| \le \frac{l}{2}$. This design choice mitigates redundant exploration of excessively large regions that do not yield optimal solutions, thereby reducing the dimensionality of the observation space and limiting unproductive search trajectories.

\subsubsection{Symmetry Cancellation}
\label{sec:symmetry}

Symmetrical structures within the protein folding environment can lead to ambiguity in Q-learning algorithms, as they may produce identical Q-values for different symmetrical actions, particularly during the initial moves. This redundancy not only increases the computational burden but also impedes the agent's ability to discern optimal folding pathways with the confusing Q-values. 

\begin{figure}[htbp] % Use figure* for a figure spanning two columns
    \centering
    \begin{subfigure}[t]{0.3\textwidth}
        \centering
        \includegraphics[width=\textwidth]{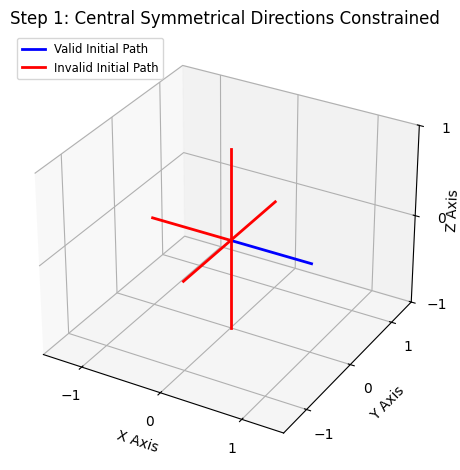}
        \label{fig:subfig1}
    \end{subfigure}
    \hfill
    \begin{subfigure}[t]{0.3\textwidth}
        \centering
        \includegraphics[width=\textwidth]{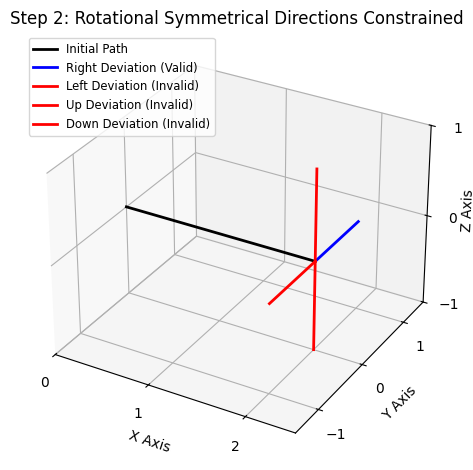}
        \label{fig:subfig2}
    \end{subfigure}
    \hfill
    \begin{subfigure}[t]{0.3\textwidth}
        \centering
        \includegraphics[width=\textwidth]{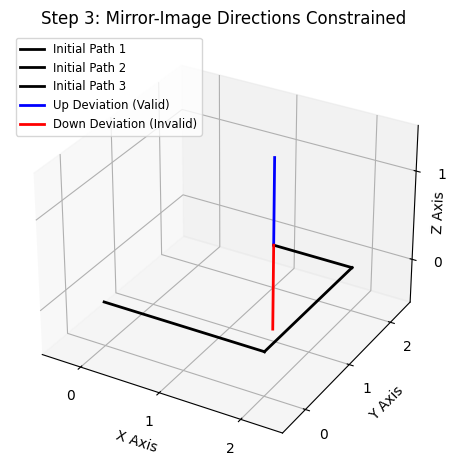}
        \label{fig:subfig3}
    \end{subfigure}
    \caption{\textbf{Three steps for canceling symmetrical structures.} Step 1: Eliminate the first action, canceling 6 central symmetrical structures around (0,0,0). Step 2: Eliminate the first non-forward action, canceling 4 rotational symmetrical structures around the x-axis. Step 3: Eliminate the first up-or-down action, canceling 2 mirror-image structures around the xy-plane.}
    \label{fig:sym_cancel}
\end{figure}

To mitigate this issue, a symmetry-breaking mechanism has been implemented by constraining the first three deviations shown in Figure~\ref{fig:sym_cancel}. The symmetry breaking is achieved through the following key constraints:

\textbf{Fixed Initial Action.} The first amino acid is positioned at coordinate $(0, 0, 0)$ and the second at $(1, 0, 0)$. This fixed placement defines the initial direction vector as $(1, 0, 0)$, canceling central symmetry structures around the coordinate origin, as well as establishing the initial direction.

\textbf{Constrained First Deviation Action.} The first deviation, i.e., any action other than forward from the initial direction is restricted to the "right" direction. This constraint cancels rotational symmetry structures around the initial axis (as the first action fixed, the x-axis). 

\textbf{Constrained First Vertical Deviation Action.} Similarly, the first vertical deviation action is restricted to the "up" direction. This constraint cancels mirror-image structures about the initial plane (as the first two constrains defined, the x-y plane).

\subsubsection{Invalid Move Avoidance}
\label{sec:invalid_move}

\begin{figure}
    \centering
    \includegraphics[width=1\linewidth]{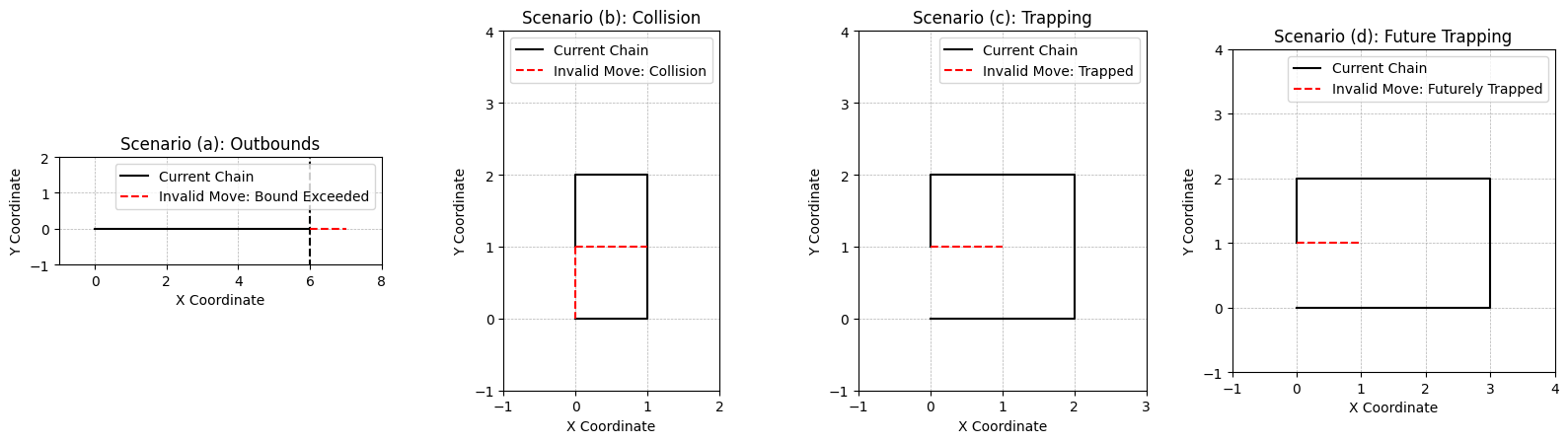}
    \caption{\textbf{Invalid move scenarios for a sequence of length 13 on a 2D lattice.} In scenario (a), the depicted move exceeds the outbound limit of $\frac{length}{2} = 6.5$. In scenario (b), an overlap occurs at the amino acid positions $(0,0)$ or $(1,1)$ for the two depicted moves. In scenario (c), no further moves can be made without resulting in a collision after the depicted move. In scenario (d), the depicted move is currently valid and not trapped, but prevents the successful placement of all 13 amino acids in subsequent steps.}
    \label{fig:invalid_moves}
\end{figure}

To ensure the validity of the folding process, the environment employs an invalid-move avoidance system. Figure~\ref{fig:invalid_moves} illustrates four examples of invalid move scenarios for a sequence of length 13. For simplicity and clarity, these examples are depicted using a two-dimensional lattice.

If an action results in an overlap with an already placed amino acid, the move is deemed invalid, and the environment terminates the episode with a reward of zero, provided the folding process is not yet complete. 

The environment checks for trapping scenarios—situations where no valid actions remain to continue the folding process. In such cases, the episode is also terminated with a zero reward, signaling the agent to avoid configurations that lead to dead-ends.

Any action that attempts to place an amino acid outside the bounding region is considered invalid. This restriction ensures that positions whose coordinates exceed $\pm \frac{l}{2}$ in any dimension are disallowed, thereby preventing coordinates from straying beyond the intended search space.

To enhance the robustness of the environment, an auxiliary function utilizing a depth-first search (DFS) algorithm is integrated to evaluate the feasibility of completing the folding process from any given state. The DFS assesses the validity of every move, as well as the availability to place all remaining amino acids without encountering trapping scenarios in the future. 

After performing the validity checks, including the symmetry-canceling constraints detailed in Section \ref{sec:symmetry}, a validity mask for each action is generated and utilized as an output mask within the agent's neural network.

\subsubsection{Reward Mechanism}

The primary reward mechanism is based on the formation of hydrophobic (H-H) contacts between non-sequential 'H' residues. Upon successful placement of all amino acids, the agent receives a reward proportional to the number of favorable H-H contacts. A reward of zero is returned after an unfinished episode or after any invalid action described in 2.1.4.

\subsection{Agent Training} 

In this section, we detail our methodology for training a Deep Q-Network (DQN)\cite{mnih2015human}-based agent, enhanced with a Transformer architecture\cite{vaswani2017attention}, dueling\cite{wang2016dueling} and double\cite{van2016deep} estimators, and prioritized replay memory\cite{horgan2018distributed}. At the core of a DQN is the concept of an action-value function, or Q-value, denoted by \(Q(s, a)\). This function expresses the expected discounted return when taking action \(a\) in state \(s\). A DQN leverages a deep neural network to approximate \(Q(s, a)\), enabling the agent to estimate optimal actions in high-dimensional state spaces.

The overall workflow can be summarized as follows: 
\begin{enumerate}
    \item The agent observes the current state (i.e., partial folding configuration).
    \item It selects an action via an \(\epsilon\)-greedy policy.
    \item The environment transitions to the next state and returns the reward.
    \item This transition is stored in a prioritized replay buffer.
\end{enumerate}
After every terminated episode, the agent samples mini-batches from the replay buffer to update the Q-network by backpropagation. A separate target network is updated at fixed intervals to stabilize training\cite{van2016deep}.

To enhance the Q-value predicting network, we employ a Transformer architecture instead of the conventional Convolutional Neural Network (CNN) typically used in DQN. The Transformer introduces an additional layer for calculating \textit{attention}, which allows the model to weigh the significance of different parts of the input data dynamically. This attention mechanism is pivotal for capturing complex dependencies within the protein folding configurations. To effectively utilize the Transformer, a preprocessing step known as positional encoding is necessary. Details of the positional encoding and the Transformer-based Q-network are discussed in the subsequent sections.

\subsubsection{Positional Encoding}
Transformer architectures rely on positional encoding to handle the lack of inherent positional information in self-attention \cite{vaswani2017attention}. In natural language tasks, each position in the token sequence is mapped to a sinusoidal encoding; similarly, we embed the position of each amino acid in the HP chain into a high-dimensional space. In this work, each residue index \(i\) in the protein chain is mapped to a sinusoidal encoding defined by:
\begin{equation}
\label{eq:positional_encoding_sin_cos}
\begin{aligned}
PE(i, 2k) &= \sin\!\Bigl(\frac{i}{10000^{\frac{2k}{d_{\mathrm{model}}}}}\Bigr),\\
PE(i, 2k + 1) &= \cos\!\Bigl(\frac{i}{10000^{\frac{2k}{d_{\mathrm{model}}}}}\Bigr),
\end{aligned}
\end{equation}
where:
\begin{itemize}
    \item \(PE(i, 2k)\) and \(PE(i, 2k + 1)\) represent the sinusoidal positional encodings for the even and odd dimensions, respectively.
    \item \(i\) denotes the residue index within the protein chain.
    \item \(k\) is the dimension index.
    \item \(d_{\mathrm{model}}\) is the dimensionality of the Transformer model's embeddings.
\end{itemize}
These encodings are then added elementwise to the projection of the input tokens described in Section \ref{sec:transformer_q_network}.

\subsubsection{Transformer-Q Network}
\label{sec:transformer_q_network}
Transformers have been successful in capturing long-range dependencies in sequential data \cite{vaswani2017attention}. We therefore adopt a Transformer-based backbone to model state representations in our DQN. The implementation leverages the \textsc{PyTorch} deep learning framework, specifically the \textsc{torch.nn} module \cite{NEURIPS2019_9015}.

The first step involves representing each state \(s_t\) as a sequence of vectors \(\{\mathbf{x}_1, \mathbf{x}_2, \ldots, \mathbf{x}_L\}\) of length \(L\). Each \(\mathbf{x}_i\) contains spatial coordinates, amino acid type, and a position index:
\begin{equation}
\label{eq:transformer_input}
\mathbf{x}_{i} \;=\; \Bigl[\mathbf{coord}_i,\;\mathbf{typeEmb}(\mathrm{type}_i),\;\frac{i}{L-1}\Bigr].
\end{equation}
Here:
\begin{itemize}
    \item \(\mathbf{coord}_i \in \mathbb{R}^3\) represents the spatial coordinates of the \(i\)-th residue.
    \item \(\mathrm{typeEmb}(\mathrm{type}_i) \in \mathbb{R}^{d_{\mathrm{type}}}\) is the embedding vector corresponding to the amino acid type of the \(i\)-th residue, where \(d_{\mathrm{type}}\) is the dimensionality of the type embeddings.
    \item \(\frac{i}{L-1}\) is the position index of the residue.
\end{itemize}

Before passing \(\mathbf{x}_{i}\) to the Transformer, we add positional encodings and prepend a special classification token \(\mathbf{x}_{\texttt{[CLS]}}\). The CLS token serves as a global representation that aggregates information from the entire sequence, facilitating the extraction of a comprehensive state representation. Specifically, for a batch size \(B\) and sequence length \(L\), the CLS token \(\mathbf{x}_{\texttt{[CLS]}}\) is replicated across the batch and concatenated to the beginning of each sequence. The resulting sequence \(\mathbf{H}^{(0)}\) has the shape:
\begin{equation}
\mathbf{H}^{(0)} \in \mathbb{R}^{B \times (L + 1) \times d_{\mathrm{model}}},
\end{equation}
where:
\begin{itemize}
    \item \(B\) is the batch size.
    \item \(L\) is the sequence length of residues.
    \item \(d_{\mathrm{model}}\) is the dimensionality of the Transformer embeddings.
\end{itemize}

We employ \(N\) Transformer encoder layers, each comprising a multi-head self-attention mechanism followed by a position-wise feed-forward network. These layers process the input sequence to capture intricate dependencies and interactions among the residues, producing enriched representations at each layer. Specifically, each encoder layer transforms the input sequence \(\mathbf{H}^{(l-1)}\) into \(\mathbf{H}^{(l)}\) as follows:
\begin{equation}
\mathbf{H}^{(l)} \;=\; \mathrm{TransformerEnc}^{(l)}\!\bigl(\mathbf{H}^{(l-1)}\bigr),
\quad l \;=\; 1,\ldots,N.
\end{equation}
where:
\begin{itemize}
    \item \(l\) indexes the Transformer encoder layers.
    \item \(N\) is the total number of Transformer encoder layers employed in the network.
    \item \(\mathbf{H}^{(l)}\) is the output representation after the \(l\)-th Transformer encoder layer.
\end{itemize}

After \(N\) layers, we take the hidden state of the \(\texttt{[CLS]}\)-token, \(\mathbf{h}_{\texttt{cls}}^{(N)}\), as a global representation of the state:
\begin{equation}
\mathbf{h}_{\texttt{cls}}^{(N)} \in \mathbb{R}^{B \times d_{\mathrm{model}}}.
\end{equation}

\paragraph{Detailed Transformer Encoder Layer Flow.}
The detailed dataflow within each Transformer encoder layer is outlined below. Notably, this implementation leverages the \textsc{torch.nn.Transformer\\EncoderLayer} module for the majority of the architecture. No custom subclass or additional parameters have been defined for the Transformer encoder, ensuring a direct utilization of the \textsc{PyTorch} library's standard implementation.

Within each encoder layer, the following operations are performed:
\begin{enumerate}
    \item \textbf{Multi-Head Self-Attention:} Computes attention scores to weigh the importance of different residues.
    \item \textbf{Residual Connection and Layer Normalization:} Adds the input to the attention output and normalizes the result.
    \item \textbf{Position-Wise Feed-Forward Network (FNN):} Applies a two-layer feed-forward network with a ReLU activation.
    \item \textbf{Final Residual Connection and Layer Normalization:} Adds the attention-based representation to the FNN output and normalizes the result.
\end{enumerate}

Specifically, the computations are as follows:
\begin{equation}
\mathbf{Q} = \mathbf{H}^{(l-1)}\,W_Q,
\quad
\mathbf{K} = \mathbf{H}^{(l-1)}\,W_K,
\quad
\mathbf{V} = \mathbf{H}^{(l-1)}\,W_V.
\end{equation}
where:
\begin{itemize}
    \item \(\mathbf{Q}\), \(\mathbf{K}\), and \(\mathbf{V}\) are the query, key, and value matrices, respectively.
    \item \(W_Q \in \mathbb{R}^{d_{\mathrm{model}} \times d_k}\), \(W_K \in \mathbb{R}^{d_{\mathrm{model}} \times d_k}\), and \(W_V \in \mathbb{R}^{d_{\mathrm{model}} \times d_v}\) are learnable weight matrices for queries, keys, and values.
    \item \(d_k\) and \(d_v\) are the dimensionalities of the queries/keys and values, respectively.
\end{itemize}

Splitting into \(H\) heads, each head employs scaled dot-product attention:
\begin{equation}
\label{eq:scaled_dotprod_attn}
\mathrm{Attn}\!\bigl(\mathbf{Q}_h, \mathbf{K}_h, \mathbf{V}_h\bigr) 
\;=\;
\mathrm{softmax}\!\Bigl(\frac{\mathbf{Q}_h \,\mathbf{K}_h^\top}{\sqrt{\,d_k\,}}\Bigr)\,\mathbf{V}_h,
\end{equation}
where:
\begin{itemize}
    \item \(H\) is the number of attention heads.
    \item \(\mathbf{Q}_h\), \(\mathbf{K}_h\), and \(\mathbf{V}_h\) are the query, key, and value matrices for the \(h\)-th head.
    \item \(d_k = \frac{d_{\mathrm{model}}}{H}\) is the dimensionality of each attention head.
\end{itemize}

After computing attention for all heads, the outputs are concatenated and projected via \(W^O \in \mathbb{R}^{(H \cdot d_v) \times d_{\mathrm{model}}}\), a learnable linear transformation that ensures the final attention output matches the model’s embedding size. Subsequently, a residual connection and layer normalization (LN) are applied:
\begin{equation}
\label{eq:attn_resnorm}
\mathbf{X}_{\text{att}}^{\text{final}}
=
\mathrm{LN}\!\bigl(
\mathbf{H}^{(l-1)} + \mathbf{X}_{\text{att}}
\bigr).
\end{equation}
where:
\begin{itemize}
    \item \(\mathbf{X}_{\text{att}}\) is the output from the multi-head self-attention mechanism.
    \item \(\mathrm{LN}\) denotes layer normalization.
\end{itemize}

Next, a position-wise feed-forward network (FNN) is applied:
\begin{equation}
\mathbf{X}_{\mathrm{FNN}}
=
\max\!\Bigl(0,\;\mathbf{X}_{\text{att}}^{\text{final}} W_1 + b_1\Bigr)\,W_2 + b_2,
\end{equation}
where:
\begin{itemize}
    \item \(W_1 \in \mathbb{R}^{d_{\mathrm{model}} \times d_{\mathrm{ff}}}\) and \(W_2 \in \mathbb{R}^{d_{\mathrm{ff}} \times d_{\mathrm{model}}}\) are learnable weights for the FNN.
    \item \(b_1 \in \mathbb{R}^{d_{\mathrm{ff}}}\) and \(b_2 \in \mathbb{R}^{d_{\mathrm{model}}}\) are bias vectors.
    \item \(d_{\mathrm{ff}}\) is the dimensionality of the feed-forward network's hidden layer.
\end{itemize}

A final residual connection and layer normalization are applied:
\begin{equation}
\mathbf{H}^{(l)}
=
\mathrm{LN}\!\bigl(
\mathbf{X}_{\text{att}}^{\text{final}} + \mathbf{X}_{\mathrm{FNN}}
\bigr).
\end{equation}
where:
\begin{itemize}
    \item \(\mathbf{X}_{\text{att}}^{\text{final}}\) is the output from the attention residual connection.
    \item \(\mathbf{X}_{\mathrm{FNN}}\) is the output from the feed-forward network.
\end{itemize}

After \(N\) layers, we take the hidden state of the \(\texttt{[CLS]}\)-token, \(\mathbf{h}_{\texttt{cls}}^{(N)}\), as a global representation of the state.

\subsubsection{Dueling and Double DQN}
To further improve the performance and stability of our Q-network, we integrate the Dueling and Double DQN methods into our architecture.

\paragraph{Dueling Networks.}
Following the extraction of the global state representation from the Transformer-Q network, we employ the Dueling DQN architecture to decompose the Q-value output into two distinct components: a scalar state-value function \(V(s)\) and an advantage function \(A(s,a)\) \cite{wang2016dueling}. In this context:
\begin{itemize}
    \item \(V(s)\) estimates the value of being in state \(s\), representing the expected return from state \(s\) regardless of the action taken.
    \item \(A(s,a)\) assesses the advantage of taking action \(a\) in state \(s\), indicating how much better taking action \(a\) is compared to the average action in state \(s\).
\end{itemize}
This decomposition allows the network to separately evaluate the intrinsic value of states and the relative benefits of actions, thereby facilitating more nuanced learning.

The final Q-values are computed as:
\begin{equation}
\label{eq:dueling_dqn}
Q(s,a;\theta) \;=\; V(s;\theta) \;+\; A(s,a;\theta)
\;-\;\frac{1}{|\mathcal{A}|} \sum_{a'} A(s,a';\theta).
\end{equation}
In this equation:
\begin{itemize}
    \item \(Q(s,a;\theta)\) denotes the Q-value for state \(s\) and action \(a\), parameterized by \(\theta\), which represents the weights of the Q-network.
    \item \(V(s;\theta)\) is the state-value function estimating the value of state \(s\).
    \item \(A(s,a;\theta)\) is the advantage function estimating the relative benefit of action \(a\) in state \(s\).
    \item \(|\mathcal{A}|\) signifies the cardinality of the action space, i.e., the total number of possible actions available to the agent.
\end{itemize}
This structure enables the network to discern the value of states independently from the advantages of actions.

\paragraph{Double Q-Learning.}
In standard DQN, the target value for updating the Q-network is computed as:
\begin{equation}
\label{eq:dqn_target}
y_t \;=\; r_{t+1} \;+\; \gamma \; \max_{a'}\,Q\bigl(s_{t+1},a';\theta^{-}\bigr).
\end{equation}
Here:
\begin{itemize}
    \item \(y_t\) represents the target Q-value for the current state-action pair at time step \(t\).
    \item \(r_{t+1}\) is the immediate reward received after taking action \(a\) in state \(s_t\).
    \item \(\gamma \in [0,1)\) is the discount factor that balances the importance of immediate and future rewards.
    \item \(\theta^{-}\) denotes the parameters of the target network, a separate neural network that is periodically updated to stabilize training.
\end{itemize}

However, standard DQN is prone to overestimation biases, where Q-values can become inflated, leading to suboptimal policy learning \cite{van2016deep}. To mitigate this issue, we adopt Double DQN, which decouples the selection and evaluation of action values \cite{van2016deep}. The Double DQN target is formulated as:
\begin{equation}
\label{eq:double_dqn_target}
y_t \;=\; r_{t+1} \;+\; \gamma \;Q\!\Bigl(
s_{t+1}, 
\arg\max_{a'}\,Q\bigl(s_{t+1},a';\theta\bigr);\,
\theta^{-}
\Bigr).
\end{equation}
In this formulation:
\begin{itemize}
    \item The action selection \(\arg\max_{a'}\,Q\bigl(s_{t+1},a';\theta\bigr)\) is performed using the current policy network with parameters \(\theta\).
    \item The action evaluation \(Q\bigl(s_{t+1},a';\theta^{-}\bigr)\) leverages the target network with parameters \(\theta^{-}\).
\end{itemize}
By separating the roles of the policy and target networks in action selection and evaluation, Double DQN effectively reduces the overestimation of Q-values. This decoupling enhances the stability and reliability of the learning process, leading to more accurate policy development.

Through the integration of Dueling and Double DQN, our agent benefits from both refined state-value estimations and mitigated Q-value overestimations. These enhancements are seamlessly incorporated into the dataflow, ensuring that each component of the network contributes optimally to the agent's learning and decision-making capabilities.

%%%%%%%%%%%%%% Discuss in the result of the benefit?

\subsubsection{Prioritized Replay Memory}
\label{prm}
In the context of the HP folding model, the presence of multiple optimal solutions can result in several actions of a state having similar high Q-values. This ambiguity may confuse the agent during training. To address this challenge and encourage the agent to make more decisive actions, we incorporate Prioritized Replay Memory (PER) into our training methodology. PER allows the agent to focus more on transitions with higher temporal-difference (TD) errors, which are typically more informative and contribute significantly to learning \cite{horgan2018distributed}.

Prioritized Experience Replay (PER) replays transitions with high temporal-difference (TD) errors more frequently. We store transitions \((s,a,r,s',d)\) with priorities \(p_i\), typically the absolute TD error plus a small constant:
\begin{equation}
p_i = \lvert\delta_i\rvert + \epsilon.
\end{equation}
Here:
\begin{itemize}
    \item \(s\) and \(s'\) denote the current and next states, respectively.
    \item \(a\) represents the action taken.
    \item \(r\) is the reward received after taking action \(a\).
    \item \(d \in \{0,1\}\) is a flag indicating whether the next state \(s'\) is terminal (\(d=1\)) or non-terminal (\(d=0\)).
    \item \(\delta_i\) is the TD error for transition \(i\), calculated as the difference between the predicted Q-value and the target Q-value.
    \item \(\epsilon > 0\) is a small positive constant to ensure all transitions have a non-zero probability of being sampled.
\end{itemize}

The probability of sampling a transition \(i\) from the replay buffer is:
\begin{equation}
\label{eq:per_prob}
P(i) \;=\; \frac{p_i^\alpha}{\sum_k p_k^\alpha},
\end{equation}
where \(\alpha \in [0,1]\) controls the degree of prioritization, with \(\alpha=0\) corresponding to uniform sampling and \(\alpha=1\) corresponding to full prioritization based on TD error. To correct the bias introduced by non-uniform sampling, we apply importance-sampling weights:
\begin{equation}
\label{eq:per_isw}
w_i \;=\; \Bigl(\frac{1}{N} \cdot \frac{1}{P(i)}\Bigr)^\beta.
\end{equation}
where:
\begin{itemize}
    \item \(w_i\) is the importance-sampling weight for transition \(i\).
    \item \(N\) is the total number of transitions stored in the replay buffer.
    \item \(\beta \in [0,1]\) controls the amount of correction, with \(\beta=0\) ignoring the weights and \(\beta=1\) fully compensating for the non-uniform probabilities.
\end{itemize}
In our implementation, \(\beta\) is annealed from a smaller value to a larger value over the course of training to progressively compensate for the bias introduced by prioritized sampling.

\subsubsection{Loss and Optimizer}
We train the Q-network by sampling mini-batches of transitions \(\{(s_j,a_j,r_j,s_j',d_j)\}\) from the prioritized replay buffer as described in Section \ref{prm}. The Q-value of the current policy network is defined as:
\begin{equation}
Q_{\theta}(s_j,a_j) \;=\; \mathrm{DuelingTransformer}\!\bigl(s_j;\theta\bigr)[a_j].
\end{equation}
Here:
\begin{itemize}
    \item \(Q_{\theta}(s_j,a_j)\) is the predicted Q-value for state \(s_j\) and action \(a_j\), parameterized by \(\theta\).
    \item \(\mathrm{DuelingTransformer}\) denotes the Transformer-based Q-network incorporating the Dueling DQN architecture.
\end{itemize}

The target Q-value \(y_j\) for transition \(j\) is computed using Double DQN as previously detailed:
\begin{equation}
y_j \;=\; r_j + (1 - d_j)\,\gamma \,
Q_{\theta^-}\!\Bigl(
s_j',\, \arg\max_{a'} Q_{\theta}\!\bigl(s_j',a'\bigr)
\Bigr).
\end{equation}
In this equation:
\begin{itemize}
    \item \(y_j\) is the target Q-value for transition \(j\).
    \item \(r_j\) is the immediate reward received after taking action \(a_j\) in state \(s_j\).
    \item \(d_j \in \{0,1\}\) indicates whether the next state \(s_j'\) is terminal (\(d_j=1\)) or non-terminal (\(d_j=0\)).
    \item \(\gamma \in [0,1)\) is the discount factor that determines the importance of future rewards.
    \item \(Q_{\theta^-}\) represents the target network with parameters \(\theta^-\), which is periodically updated to match the policy network parameters \(\theta\).
    \item \(\arg\max_{a'} Q_{\theta}\!\bigl(s_j',a'\bigr)\) selects the action \(a'\) that maximizes the Q-value for the next state \(s_j'\) as predicted by the current policy network.
\end{itemize}

The temporal-difference (TD) error is defined as:
\begin{equation}
\delta_j = y_j \;-\; Q_{\theta}(s_j,a_j).
\end{equation}
where \(\delta_j\) quantifies the difference between the predicted Q-value and the target Q-value for transition \(j\). Our base loss function is the weighted mean-squared error:
\begin{equation}
\label{eq:loss_mse}
\mathcal{L}(\theta) \;=\; \sum_{j} w_j \bigl(\delta_j\bigr)^2,
\end{equation}
where:
\begin{itemize}
    \item \(\mathcal{L}(\theta)\) is the loss function to be minimized.
    \item \(w_j\) are the importance-sampling weights from Equation \eqref{eq:per_isw}.
\end{itemize}

We use the Adam optimizer \cite{kingma2014adam} to minimize the loss function. After each update, the absolute TD errors \(\lvert \delta_j\rvert\) are used to update replay priorities, ensuring that transitions with higher errors are sampled more frequently in future updates. Additionally, the target network parameters \(\theta^{-}\) are periodically synchronized to match the policy network parameters \(\theta\).

\section{Experiments and Discussion}

The training procedures were conducted on a desktop workstation with the following specifications:
\begin{itemize}
    \item \textbf{CPU}: Intel Core i9-12900K
    \item \textbf{Memory}: 2$\times$32\,GB and 2$\times$16\,GB, totaling 96\,GB of DDR4 non-ECC memory
    \item \textbf{GPU}: NVIDIA GeForce RTX 3090 Ti
\end{itemize}

Additionally, training was also performed on a supercomputer cluster equipped with NVIDIA A100 GPUs (detailed specifications are provided in \cite{wisteria_system}). All training parameters were manually selected. \textbf{It should be noted that, due to strict project deadlines:}

\begin{itemize}
    \item The parameters were not comprehensively fine-tuned and may therefore be suboptimal. 
    \item Training runs achieving the best-known results for long sequences ($N \geq 36$) were performed only once, which may compromise reproducibility. 
\end{itemize}

In the following, we present detailed benchmarks, comparisons with related studies, and further discussion.

\subsection{Benchmarks and Comparisons}
\label{sec:benchmark}

We trained our transformer-enhanced Q-network using the seven sequences listed in Table~\ref{tab:benchmarK_series}. These sequences were introduced in \cite{unger1993genetic} as benchmarks for 2D HP models and are widely employed in subsequent 3D HP model research \cite{lin2011protein, traykov2018algorithm, boumedine2021protein, rezaei2024novel, yanev2017protein}. 

Throughout the training, periodic evaluations were performed by running a complete episode with $\epsilon=0$ to assess the final reward. During both training and evaluation, the best reward was recorded and updated whenever a superior reward was identified. The coordinates (i.e., the final chain structure) were also recorded whenever the best-known reward improved. 

\begin{table}[ht]
	\centering
	\caption{\textbf{3D lattice HP benchmarks.}
		The ``Energy'' column lists the best-known values reported in \cite{boumedine2021protein}, which align with the best rewards found at the time of submission.}
	\label{tab:benchmarK_series}
	\begin{tabular}{lccc}
		\hline
		Seq. & Length & Protein Sequence & Best Known Energy \\
		\hline
		1 & 20 & (HP)$_2$PH(HP)$_2$(PH)$_2$HP(PH)$_2$ & -11 \\
		2 & 24 & H$_2$P$_2$(HP$_2$)$_6$H$_2$ & -13 \\
		3 & 25 & P$_2$HP$_2$(H$_2$P$_4$)$_3$H$_2$ & -9 \\
		4 & 36 & P(P$_2$H$_2$)$_2$P$_5$H$_5$(H$_2$P$_2$)$_2$P$_2$H(HP$_2$)$_2$ & -18 \\
		5 & 48 & P$_2$H(P$_2$H$_2$)$_2$P$_5$H$_{10}$P$_6$(H$_2$P$_2$)$_2$HP$_2$H$_5$ & -31 \\
		6 & 50 & H$_2$(PH)$_3$PH$_4$PH(P$_3$H)$_2$P$_4$(HP$_3$)$_2$HPH$_4$(PH)$_3$PH$_2$ & -34 \\
		7 & 60 & P(PH$_3$)$_2$H$_5$P$_3$H$_{10}$PHP$_3$H$_{12}$P$_4$H$_6$PH$_2$PHP & -55 \\
		\hline
	\end{tabular}
\end{table}

For Sequences~1--7 in Table~\ref{tab:benchmarK_series}, the training parameters used in this study are presented in Table~\ref{tab:model_performance}. Due to deadline constraints, these parameters have not been rigorously fine-tuned, and their scalability for different model sizes remains unclear, which may explain their apparent lack of order or systematic arrangement. Nonetheless, the parameters in Table~\ref{tab:model_performance} produced the best performance during our internal experiments.

%%%%%%%% tab:model_performance

\begin{table} % Do NOT use \begin{table*}
	\centering
	% Captions go above tables
        \caption{\textbf{Parameters Utilized in Optimal Runs.} The parameters were not scaled systematically. Attempts to scale the parameters in relation to model size did not reveal a consistent scaling rule. Nevertheless, these parameter configurations have been demonstrated to yield relatively favorable results.}
	\label{tab:model_performance} % give each table a logical label name
	
	\begin{tabular}{ccccccc} % Seven columns for data
		\\
		\hline
		Seq & Reached Best & Episodes & Learning Rate & Batch Size & dim\_model & num\_layer\\
		\hline
		1 & -11 & 80,000 & 5.00E-04 & 512 & 64 & 1\\
		2 & -13 & 100,000 & 5.00E-05 & 256 & 256 & 2\\
		3 & -9 & 100,000 & 5.00E-04 & 1024 & 256 & 2\\
		4 & -18 & 100,000 & 5.00E-04 & 1024 & 256 & 2\\
		5 & -29 & 300,000 & 5.00E-04 & 2048 & 256 & 3\\
		6 & -29 & 400,000 & 5.00E-04 & 2048 & 256 & 3\\
		7 & -49 & 300,000 & 2.00E-05 & 256 & 512 & 3\\
		\hline
	\end{tabular}
\end{table}

Table~\ref{tab:algorithm_comparison} compares the best energy values achieved in our work with those of related algorithms. The proposed model is labeled ``Attn-DRL.'' As shown in Table~\ref{tab:algorithm_comparison}, the proposed method attained the best-known values for Sequences~1--4. For Sequences~5--7, our model achieved superior values compared to five state-of-the-art approaches (ACO, GA, EMC, PERM, and TRL-PSSPSO) \cite{rezaei2024novel}---which do not employ neural networks---although it did not reach the overall best-known values. Notably, prior to the update reported by EHA \cite{yanev2017protein} in 2017, the best-known values for Sequences~5--7 equaled those achieved in our experiments. While several studies address the 2D HP model using neural networks and/or DRL architectures, we were unable to find analogous work focusing on the 3D HP model. 

\begin{table}[ht]
    \centering
    \caption{\textbf{Comparison of performance across algorithms.}
         ACO, GA, EMC, PERM, and TRL-PSSPSO results are taken from \cite{rezaei2024novel}. The EHA approach from \cite{yanev2017protein} further improved upon earlier best-known values for Sequences~5--7 in 2017. Subsequently, HCSA \cite{boumedine2021protein} updated the best-known value for Sequence~6 in 2021.}
    \label{tab:algorithm_comparison}
    \begin{tabular}{lcccccccc}
        \hline
        Seq. & ACO & GA & EMC & PERM & TRL-PSSPSO & \textbf{Attn-DRL} & HCSA & EHA \\
        \hline
        1 & $-9$ & $-9$ & $-9$ & $-9$ & $-11$ & $\textbf{-11}$ & $-11$ & $-11$ \\
        2 & $-9$ & $-9$ & $-9$ & $-9$ & $-11$ & $\textbf{-13}$ & $-13$ & $-13$ \\
        3 & $-8$ & $-8$ & $-8$ & $-8$ & $-9$ &  $\textbf{-9}$ & $-9$ & $-9$ \\
        4 & $-14$ & $-14$ & $-14$ & $-14$ & $-16$ & $\textbf{-18}$ & $-18$ & $-18$ \\
        5 & $-23$ & $-23$ & $-23$ & $-23$ & $-25$ & $\textbf{-29}$ & $-31$ & $-31$ \\
        6 & $-21$ & $-21$ & $-21$ & $-21$ & $-24$ & $\textbf{-29}$ & $-31$ & $-28$ \\
        7 & $-36$ & $-34$ & $-35$ & $-36$ & $-48$ & $\textbf{-49}$ & $-54$ & $-55$ \\
        \hline
    \end{tabular}
\end{table}

\subsection{Training Process}
\label{sec:training_process}

\begin{figure}[htbp]
    \centering
    \begin{subfigure}[t]{0.49\textwidth}
        \centering
        \includegraphics[width=\textwidth]{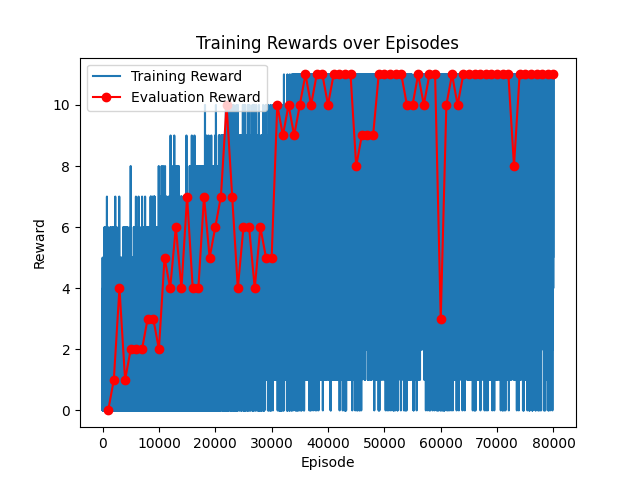}
        \subcaption{Training Curve of Sequence 1}
    \end{subfigure}
    \hfill
    \begin{subfigure}[t]{0.49\textwidth}
        \centering
        \includegraphics[width=\textwidth]{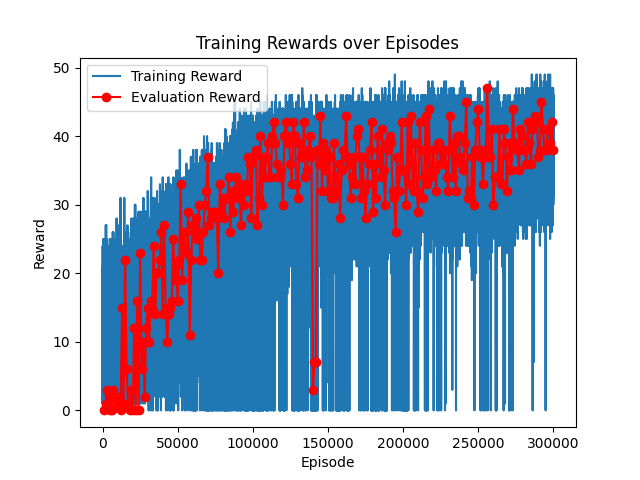}
        \subcaption{Training Curve of Sequence 7}
    \end{subfigure}
    \caption{\textbf{Training Curves for Sequences 1 and 7.} The reward is assigned as a positive value equivalent to the absolute value of the energy value. In (a), the evaluation for Sequence 1 converged to an optimal value of -11 but subsequently degraded after convergence. In (b), the evaluation metric for Sequence 7 failed to stabilize at any specific value, and no evaluated model attained the optimal value of -49 identified during training.}

    \label{fig:training_curves}
\end{figure}

During the training process, we observed that periodic evaluations (as described in Section \ref{sec:benchmark}) yielded significantly suboptimal results compared to the rewards obtained from the training simulation process. Figure~\ref{fig:training_curves} illustrates examples of evaluation results alongside training simulation rewards for two sequences. As depicted in Figure~\ref{fig:training_curves}, the evaluation results either degraded multiple times after reaching the best-found value, or completely failed to converge to any stable value.

Experience suggests that such convergence issues are typically associated with insufficient parameter fine-tuning. However, despite conducting experiments that individually adjusted parameters—including learning rate, model architecture, batch size, replay memory size, and number of training episodes—while controlling for other variables, the problem persisted until the submission of this report.

\begin{figure}[htbp]
    \centering
    \begin{subfigure}[t]{0.49\textwidth}
        \centering
        \includegraphics[trim=1cm 1cm 1cm 2cm, clip, width=\textwidth]{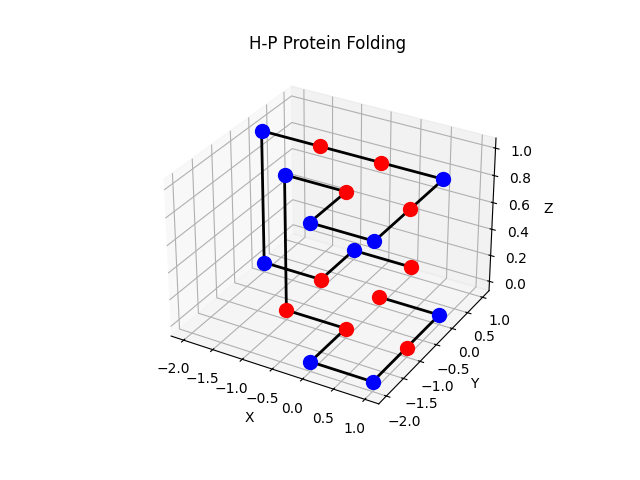}
        \subcaption{Training episode No.\ 50933}
    \end{subfigure}
    \hfill
    \begin{subfigure}[t]{0.49\textwidth}
        \centering
        \includegraphics[trim=1cm 1cm 1cm 2cm, clip, width=\textwidth]{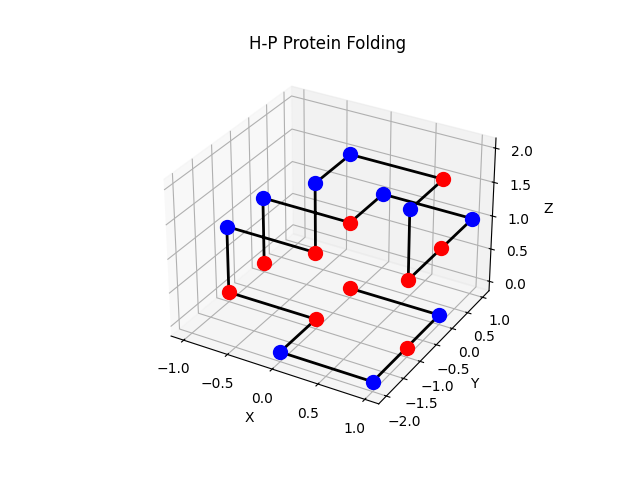}
        \subcaption{Training episode No.\ 53922}
    \end{subfigure}
    \hfill
    \begin{subfigure}[t]{0.49\textwidth}
        \centering
        \includegraphics[trim=1cm 1cm 1cm 2cm, clip, width=\textwidth]{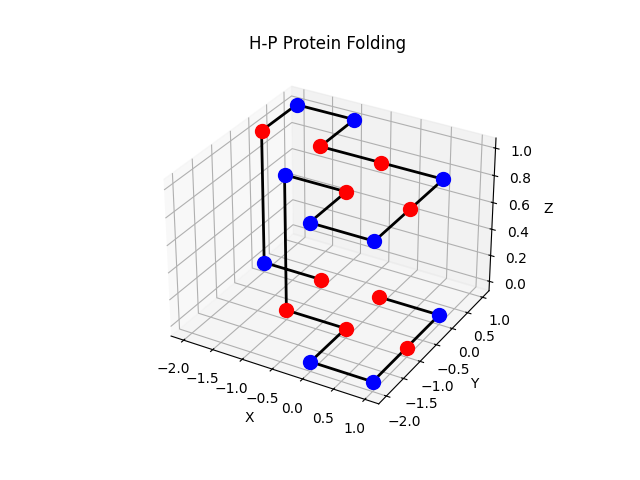}
        \subcaption{Training episode No.\ 62007}
    \end{subfigure}
    \hfill
    \begin{subfigure}[t]{0.49\textwidth}
        \centering
        \includegraphics[trim=1cm 1cm 1cm 2cm, clip, width=\textwidth]{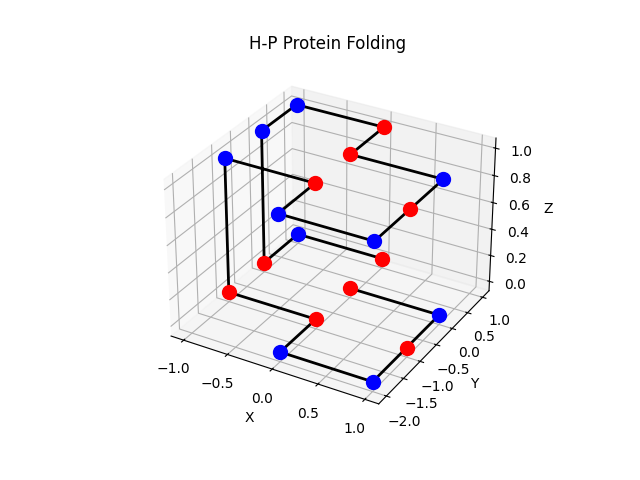}
        \subcaption{Training episode No.\ 63917}
    \end{subfigure}
    \caption{\textbf{Optimal structures identified in Sequence~1.} All structures were found during simulation episodes. Structure~(b) diverges from the others at the 9th action, while (d) differs from (a) and (c) beginning at the 14th action. Further differences occur at the 15th action between (c) and (d).}
    \label{fig:dif_structures}
\end{figure}

We also observed that the optimal solution is not unique for many sequences. Figure~\ref{fig:dif_structures} shows examples of different optimal solutions found for Sequence~1. During the same training run, 40 distinct optimal solutions were identified with an energy of $-11$, including the four structures depicted in Figure~\ref{fig:dif_structures}. In the SAW simulation, different solutions can diverge quite early, for example, the structure labeled (b) in Figure~\ref{fig:dif_structures} differs from the others starting at the 9th action.

Because the training procedure exploits the Q-values associated with each action, multiple optimal solutions with comparable Q-values but varying action paths can introduce ambiguity. Once an optimal solution is found, the policy may still be encouraged to explore alternative routes from earlier states, which could explain the occasional performance degradation noted above.

\subsection{Further Improvements}

By the submission deadline, only preliminary testing of the designed attention-based deep Q-network was feasible, and the number of experiments conducted was insufficient for ensuring reproducibility or optimizing parameters. Consequently, the training parameters may be suboptimal, and the convergence issues remain unresolved. Nonetheless, a competitive performance level was achieved using the prototype model. Further enhancements to this model could yield improved results and render transformer-based models practically effective for addressing the HP protein folding problem. This section discusses potential avenues for significant improvements in this work.

\subsubsection{Parameters fine tuning}

In this study, identifying optimal parameters for the designed network proved challenging. Given the absence of prior research applying attention-based models to the HP model, parameters from another deep reinforcement learning (DRL) study utilizing non-attention networks \cite{yang2023applying} were adopted as a starting point. However, it was discovered that attention-based models were highly sensitive to parameter settings, and optimal parameters may differ substantially. Specifically, the learning rates tested needed to be significantly lower (by a factor of 10 to 50, scaled with the batch size according to \cite{goyal2017accurate}) in larger models compared to those used in \cite{yang2023applying}.

As illustrated in Table~\ref{tab:model_performance}, the parameters were not systematically tuned, and the scaling relationships appeared random. Moreover, the convergence issues remained unresolved, indicating a substantial potential for enhancing data efficiency through further parameter fine-tuning.

\subsubsection{Advanced exploration strategy}

The epsilon-greedy strategy assumes that tasks require more exploration during the initial stages, eventually stabilizing to exploitative policies. However, as described in Section \ref{sec:training_process}, the presence of multiple optimal solutions can lead to ambiguous Q-values in later training stages. This ambiguity may incentivize the agent to explore inadequately explored alternative routes from earlier states, even if such exploration is not beneficial given the remaining training episodes and the remaining epsilon budget (i.e., the extent to which the policy permits future exploration). Although there was insufficient time to thoroughly investigate the Q-value dynamics, this phenomenon is considered a plausible cause of the observed convergence issues.

To address this problem, advanced exploration strategies may prove effective. Instead of employing a globally decreasing epsilon, implementing a per-state epsilon that adapts to each state could allow the agent to continue exploring promising yet under-explored states. This approach would optimize the utilization of the total training episode budget by dynamically adjusting exploration incentives based on state-specific information.

Alternatively, replacing the epsilon-greedy strategy with other exploration heuristics may enhance performance. An elegant solution is the use of Noisy Nets \cite{fortunato2018noisy}, introduced by Fortunato et al. in 2018, which injects noise directly into the network parameters. Over the course of training, the network can learn optimal noise levels within its layers, potentially mitigating task-specific ambiguities in Q-value outputs.

\section{Conclusion}

This study presented a Transformer-enhanced Deep Q-Network approach for solving the 3D H-P protein folding problem. Our method integrates standard DQN enhancements, such as dueling networks, double Q-learning, and prioritized replay, and further introduces symmetry-breaking constraints and an auxiliary feasibility check. Applied to established benchmark sequences, the proposed approach matches several known best solutions, especially for moderate-length chains.

Despite this progress, our experiments revealed sensitivities to hyperparameter settings and difficulties in maintaining stable policies, particularly when multiple equally optimal solutions coexist. These observations highlight the importance of systematic parameter tuning and advanced exploration strategies—such as per-state adaptive epsilon or Noisy Networks—to counteract the ambiguity arising from multiple optimal paths. 

Going forward, further refinements to the Transformer-based architecture, along with more extensive training protocols, could bridge remaining performance gaps on longer sequences and unlock the full potential of attention-based methods for protein structure prediction under the H-P model. 

Further research is also required to clarify the role of attention-based architectures in this domain. Given the time constraints of this study, we were unable to conduct rigorous ablation experiments or controlling comparisons to isolate the specific contribution of transformer layers within our DRL framework. Moreover, no prior work on DRL for the 3D cubic-lattice HP model was identified at the time of submission, leaving the exact role of the transformer component underexplored. 

Nonetheless, preliminary evidence indicates that transformer layers may substantially improve data efficiency relative to conventional CNN-based approaches. For instance, Yang et al.\cite{yang2023applying} required 500,000 trials of a non-attention CNN network to solve Sequence~4 in a 2D square lattice, whereas our attention-based model required only 100,000 episodes to solve the 3D cubic-lattice variant of the same sequence—despite its inherently larger solution space. Although differences in training processes (e.g., batch size and update frequency) limit the direct comparability of these results, our findings suggest that transformers can enhance performance and reduce the number of episodes needed for convergence. Future investigations are warranted to systematically control for these architectural and training variations in order to more precisely elucidate the transformer's impact. In particular, applying attention-based layers to the 2D square-lattice HP model, and comparing performance with existing RL studies, would offer further insights into how and why 
transformer mechanisms confer advantages in protein folding tasks.

In summary, this work highlights the significant potential of Transformer-based deep reinforcement learning methods to advance protein folding research, laying the groundwork for future innovations in computational biology.

%%%%%%%%%%%%%%%% REFERENCES %%%%%%%%%%%%%%%

\clearpage % Clear all remaining figures and tables then start a new page

% The list of references goes after the main text and before the acknowledgements
% When preparing an initial submission, we recommend you use BibTeX, like this:
%
\bibliography{science_template} % for a file named science_template.bib
\bibliographystyle{sciencemag}

\section*{Acknowledgments}

We thank Prof. Iba for granting us access to the Wisteria supercomputer, extending the report deadline, reviewing our methodology before the final training, and inviting us to have some fun in the bounenkai at Iba's lab. We also thank Fujita, Surya, and other members of Iba's lab for their assistance in accessing the supercomputer.

We acknowledge the assistance of OpenAI's GPT model in improving the clarity and quality of the English writing and providing suggestions for coding implementations.

\paragraph*{Data and Materials Availability:}

The final code for producing the best results, along with the output folder containing the runs that produced the best results for Sequence \ref{tab:benchmarK_series}, is available at: \burl{https://github.com/SakuraHoshizaki/Transformer_HP_3d/tree/main}.

\end{document}